\documentclass[runningheads]{llncs}
\usepackage[T1]{fontenc}
\usepackage{graphicx}
\usepackage{latexsym}
\usepackage{makecell}

\usepackage{CJKutf8}

\begin{document}

\begin{CJK*}{UTF8}{gbsn}

\title{Overview of the PromptCBLUE Shared Task in CHIP2023}
%
%
\author{Wei Zhu\inst{1} \and
Xiaoling Wang\inst{1} \thanks{Corresponding author. Email: \email{xlwang@cs.ecnu.edu.cn}. } \and
Mosha Chen\inst{2} \and
Buzhou Tang\inst{3} \\
}
\authorrunning{Zhu et al.}
%
\institute{East China Normal University, Shanghai, China \\
\and
Holoflow Digital Technology, Hangzhou, China \\
\and
Harbin Institute of Technology, Shenzhen, China \\
}

\maketitle              
\begin{abstract}

This paper presents an overview of the PromptCBLUE shared task\footnote{\url{http://cips-chip.org.cn/2023/eval1}} held in the CHIP-2023 Conference. This shared task reformualtes the CBLUE benchmark, and provide a good testbed for Chinese open-domain or medical-domain large language models (LLMs) in general medical natural language processing. Two different tracks are held: (a) prompt tuning track, investigating the multitask prompt tuning of LLMs, (b) probing the in-context learning capabilities of open-sourced LLMs. Many teams from both the industry and academia participated in the shared tasks, and the top teams achieved amazing test results. This paper describes the tasks, the datasets, evaluation metrics, and the top systems for both tasks. Finally, the paper summarizes the techniques and results of the evaluation of the various approaches explored by the participating teams.

\end{abstract}

\section{Introduction}

In 2023, the launch of large language models like ChatGPT, gpt-4, Claude, Bard have taken the world by surprise. People are amazed by their general capabilities: (a) universal NLP task solver \cite{2023arXiv230206476Q}. (b) powerful chatting capabilities. (c) following human instructions \cite{ouyang2022training,Cui2023UltraFeedbackBL}. (d) reasoning and planning capabilities \cite{Wei2022ChainOT}. Recent works have shown that these powerful LLMs have expert level knowledge in the medical domain \cite{singhal2023large}. Although these models are powerful, they are proprietary, and can not be deployed locally, preventing many applications that have serious privacy concerns \cite{Yao2023ASO}. Recently, more and more powerful open-sourced LLMs are being released \cite{Liu2023LLM360TF}, making it more and more convenient for developers to train or fine-tune a local LLM for application developments.

\begin{figure*}[h]
\begin{center}
\includegraphics[width=0.98\textwidth]{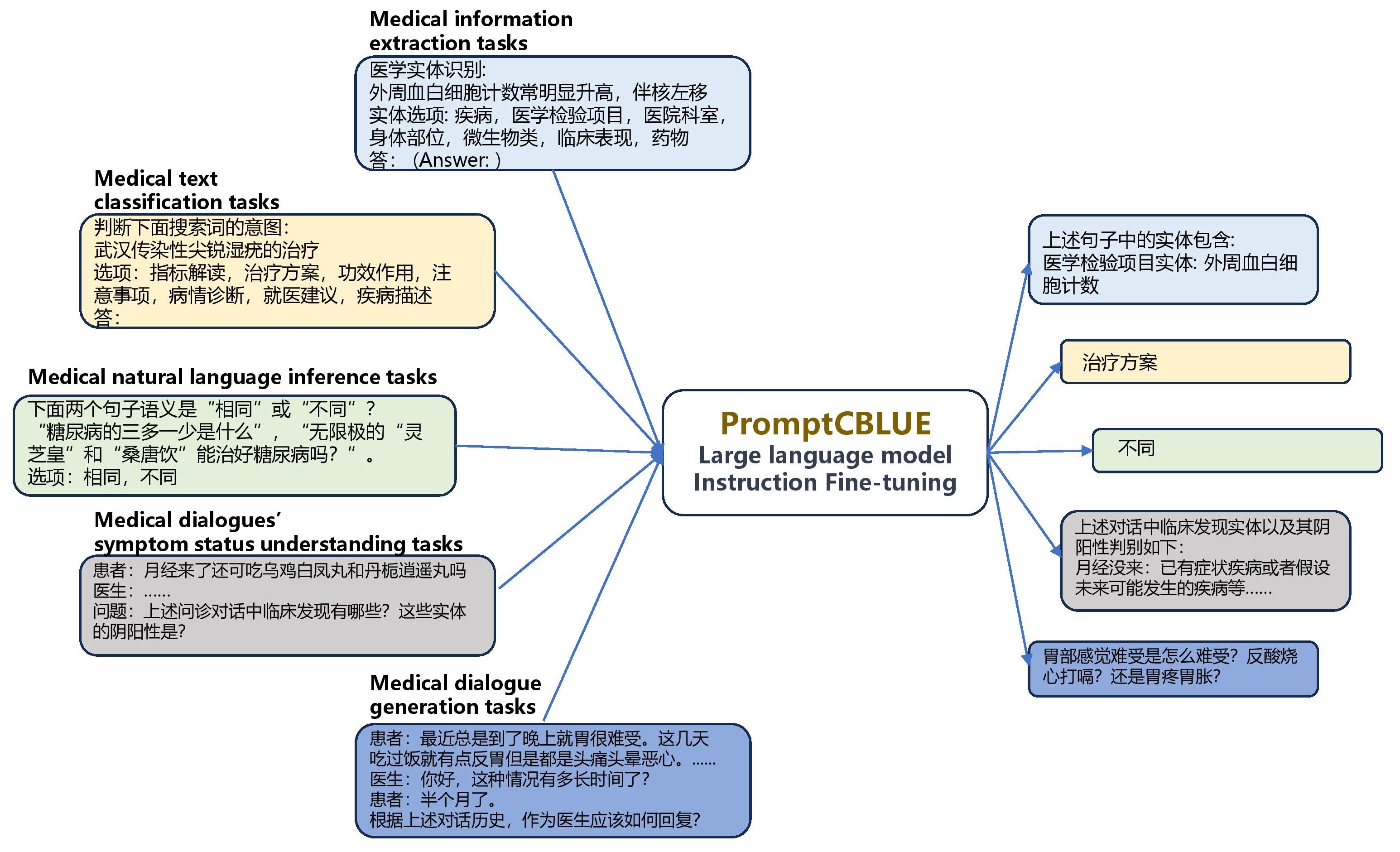}
\end{center}
\caption{We introduce PromptCBLUE, a large-scale instruction tuning benchmark for Chinese medical LLMs, which converts different types of medical natural language processing tasks into a unified prompt-response generation task. PromptCBLUE consists of five cohorts of 18 tasks, which cover a variety of medical applications. }
\label{fig:promptcblue}
\end{figure*}

In order to evaluate the multitask capabilities of the recent open-sourced Chinese LLMs in medical NLP tasks, we re-build the CBLUE benchmark \cite{zhang-etal-2022-cblue} into PromptCBLUE \cite{PromptCBLUE}, a large-scale multi-task prompt tuning benchmark dataset (see Figure \ref{fig:promptcblue}). This benchmark is one of first and largest prompt tuning benchmark for medical LLMs in Chinese. The PromptCBLUE shared task is held in the 2023 China Health Information Processing Conference, with the help of the conference committees. The shared task evaluate Chinese LLMs in two aspects, corresponding to two tracks: (a) the Parameter-efficient Fine-tuning (PEFT) Track, and (b) the In-Context Learning (ICL) Track.

The PEFT track is closely related to the recent trend of PEFT in the LLM research. Despite LLMs becoming general task solvers, fine-tuning still plays a vital role in efficient LLM inference and controlling the style of the LLMs' generated contents.\footnote{Recently, OpenAI also released the fine-tuning API for GPT-3.5-turbo. See blog post: \url{https://openai.com/blog/gpt-3-5-turbo-fine-tuning-and-api-updates}.} Fine-tuning such large models by full parameters is prohibitive since it requires a large amount of GPU memory and computations. Thus, parameter-efficient fine-tuning (PEFT) \cite{Zhang2023LearnedAA,2023arXiv230318223Z} has raised much attention in the research field since in PEFT, the tunable parameters are often less than 1\% of the LLMs and the computation costs will be significantly decreased. In the PEFT Track, participants are asked to fine-tune a given open-sourced Chinese LLM with a single PEFT module for all the 18 sub-tasks of PromptCBLUE, while keeping the backbone LLM unchanged. This track intends to challenge the participants to come up with novel PEFT modules or novel multi-task training methods.


With the recent advancements in scaling up model parameters, large language models (LLMs) showcase promising results on a variety of few-shot tasks through in-context learning (ICL), where the model is expected to directly generate the output of the test sample without updating parameters. This is
achieved by conditioning on a manually designed prompt consisting of an optional task description and a few demonstration examples \cite{radford2019language}. Since then, many efforts have been made on ICL \cite{Li2023UnifiedDR}. In the ICL Track of , participants are asked to push the limit of ICL for medium sized (6B, 7B or 13B parameters) open-sourced LLMs. The backbone LLMs are freezed and they are not allowed to introduce any additional parameters to the backbone models. However, they can improve the LLMs' performance by designing better prompts, especially select proper demonstrations for any given samples. A BERT base model or any model with a similar size is allowed to facilitate the demonstration selection. 

The PromptCBLUE shared task has raised much attention in both the industry and academia. A total of 600 teams have participated in either track of the shared task, and 48 teams have submitted predictions for the second round. In this paper, we will review the shared task, the winning teams and their methodologies, and discuss future research directions.

\section{Related Work}

\subsection{Medical natural language processing}

The developments in neural networks and natural language processing has advanced the field of medical natural language processing (MedNLP) \cite{2021arXiv211015803Z,hahn2020medical,zhu-etal-2021-discovering,Zhou2019AnalysisOT,Zhu2021pahtnlpM,Zhu2023BADGESU,Zheng2023CandidateSF}. In the pre-BERT era, firstly, RNNs like LSTM/GRU are used for processing sequential medical data such as text and speech \cite{beeksma2019predicting}. Convolutional networks are also used for medical text classificaiton \cite{hughes2017medical,Zhu2021AutoNLUAS}. The techniques of Graph neural networks are also explored for diagnose recommendations \cite{li2020graph}. In this period, many different model architectures are specially designed for better performances on a specific MedNLP task \cite{zhu-etal-2021-discovering,autotrans,Zhang2021AutomaticSN}. Since BERT \cite{devlin2018bert}, the pretrained language models (PLMs) become the deafult solution for MedNLP. In this stage, researcher becomes less interested in modifying the model architecture, but instead trying to pretrain or further pretrain a PLM from the open domain to the medical domain \cite{guo2021global,zhu-2021-mvp,pubmedbert,Zhu2021MVPBERTMP,Wang2023MultitaskEL}.

With the wide study of LLMs, the field of MedNLP is also being revolutionized. There are already works on adapting LLM backbones to the medical domain question answering \cite{zhu2023ChatMed}. And \cite{PromptCBLUE} propose PromptCBLUE, a prompt learning based benchmark dataset for examing the LLMs' ability in MedNLP tasks. This work investigates the capabilities of both the commercial LLMs like ChatGPT, and the open-sourced ones by employing the PromptCBLUE as the testbed, providing a deeper understanding of LLMs for future MedNLP research.

\subsection{Parameter-efficient Fine-tuning} 

In this subsection, we review currently the most popular PEFT methods. 

\noindent\textbf{Adapter-based tuning.} \quad One of the most important research lines of PEFT is adapter-based tuning. Adapter \cite{houlsby2019parameter} inserts adapter modules with bottleneck architecture between every consecutive Transformer \cite{Vaswani2017AttentionIA} sublayers. AdapterFusion \cite{pfeiffer-etal-2021-adapterfusion} only inserts sequential adapters after the feed-forward module. Adapter-based tuning methods have comparable results with model tuning when only tuning a fraction of the backbone model's parameter number. Due to their strong performance, a branch of literature has investigated the architecture of adapters in search of further improvements. \cite{He2021TowardsAU} analyze a wide range of PETuning methods and show that they are essentially equivalent. They also propose the general architecture of PEFT, and derive the Parallel Adapter which connects the adapter modules in parallel to the self-attention and MLP modules in the Transformer block. AdapterDrop \cite{Rckl2020AdapterDropOT} investigates the efficiency of removing adapters from lower layers. Adaptive adapters \cite{Moosavi2022AdaptableA} investigate the activation functions of adapters and propose to learn the activation functions of adapters via optimizing the parameters of rational functions as a part of the model parameters. Compacter \cite{Mahabadi2021CompacterEL} uses low-rank parameterized hypercomplex multiplication \cite{Le2021ParameterizedHG} to compress adapters' tunable parameters. LST \cite{Sung2022LSTLS} improves the memory efficiency by forming the adapters as a ladder along stacked Transformer blocks, and it enhances the adapter module by adding a self-attention module to its bottleneck architecture. \cite{Sung2022LSTLS,Jie2022ConvolutionalBA,Zhang2023FastNERSU} try to add different encoding operations, like self-attention operations and convolutions between the bottleneck structure of adapters, and achieve better performances. Learned-Adapter \cite{Zhang2023LearnedAA} builds upon the above adapter-based methods and enhance the performance of adapter tuning by automatically learning better architectures for adapters.

\noindent\textbf{Prompt tuning methods} \quad Prompt tuning \cite{lester2021power} and P-tuning \cite{Liu2022PTuningPT} insert a soft prompt to word embeddings only, and can achieve competitive results when applied to supersized PTMs. Prefix-tuning \cite{Li2021PrefixTuningOC} and P-tuning v2 \cite{Liu2021PTuningVP} insert prompts to every hidden layer of PTM. IDPG \cite{Wu2022IDPGAI} uses the prompt generator with parameterized hypercomplex multiplication \cite{Le2021ParameterizedHG} to generate a soft prompt for every instance. LPT \cite{Liu2022LatePT} improves upon IDPG by selecting an intermediate layer to start inserting prompts. SPT \cite{zhu-tan-2023-spt} designs a mechanism to automatically decide which layers to insert new instance-aware soft prompts.

\noindent\textbf{Literature for the LoRA methods} \quad Since LoRA is the most popular PEFT method in the era of large language models, there are many works that are orthogonal to AdaLoRA, SoRA and our work that are devoted to improve LoRA on many different aspects. QLoRA \cite{qlora} proposes a novel quantization method that can significantly reduce the memory consumptions of LLMs during LoRA fine-tuning. LoRA-FA \cite{Zhang2023LoRAFAML} freezes parts of the randomly initialized LoRA matrices. (d) VERA \cite{Kopiczko2023VeRAVR} investigate whether one could froze the randomly initialized LoRA matrices and only learns a set of scaling vectors. Tying LoRA matrices across layers are also investigated by VERA.

\subsection{In-context learning}

GPT-3 \cite{radford2019language}, the OpenAI's former SOTA LLM, raise the attention of the research field to a novel research direction: in-context learning (ICL). In their paper, they have showcased that GPT-3, a self-supervised pretrained model, can immediately master a new task it is never trained on by reading a manually designed prompt consisting of an optional task description and a few demonstration examples. Since then, a branch of literature has been devoted to investigate different aspects of ICL. A series of theoretical analysis attempted to understand why ICL works \cite{Kim2022GroundTruthLM,Min2022RethinkingTR,Dai2022WhyCG}. \cite{Dai2022WhyCG} explains language models as meta-optimizers and understand in-context learning as implicit fine-tuning. They prove that GPT first produces meta-gradients according to the demonstration examples, and then these meta-gradients are applied to the original GPT to build an ICL model. Since performance of ICL has been shown to be highly sensitive to the selection of demonstration examples \cite{Li2023UnifiedDR}. \cite{Rubin2021LearningTR} proposed to learn to retrieve demonstration examples. \cite{Li2023UnifiedDR} proposes a series of techniques to enhance the performance of demonstration selection. \cite{levy-etal-2023-diverse} selected
diverse demonstrations to improve in-context compositional generalization. More recent studies have explored leveraging the output distributions of language models to select few-shot demonstrations \cite{li-qiu-2023-finding}. \cite{Qin2023InContextLW} iteratively selects examples that
are diverse but still strongly correlated with the test sample as ICL demonstrations.

\section{Overview of PromptCBLUE}

\subsection{overview}

Built upon the CBLUE benchmark \cite{zhang-etal-2022-cblue}, we created an extensive multi-task test suite in the medical domain for the LLMs that supports Chinese. The tasks in PromptCBLUE can be divided into the following groups:

\begin{itemize}
\item \textbf{Medical information extraction}, including: (a) CMeEE-V2 \cite{Zan2020BuildingAP}, a medical named entity recognition task. (b) CMeIE \cite{Guan2020CMeIECA}, a medical triple extraction task. (c) CHIP-CDEE, which asks models to extract clinical finding events; (d) CHIP-CDN, which asks one to map the diagnosis descriptions to standard ICD-10\footnote{
https://www.whofic.nl/familie-van-internationale-classificaties/referentie-classificaties/icd-10} disease terms. (e) IMCS-V2-NER for extracting medical entities in the dialogues from the medical dialogue datasets IMCS-V2 \cite{Chen2022ABF}. (f) Text2DT, which is a complex medical information extraction task proposed by \cite{text2dt_shared_task}. This novel task asks a model to extract medical decision processes in the format of binary trees from the unstructured medical text. (g) CmedCausal \cite{cmedcausal}, which asks a model to extract triplets with causal relations from unstructured medical documents. 
\item \textbf{Medical text classification}, which includes: (a) CHIP-CTC \cite{Zong2021SemanticCO}, the classification task for Chinese eligibility criteria in clinical trials. (b) KUAKE-QIC, which classifies online medical queries. (c) IMCS-V2-DAC for medical intent classification of medical dialogues. 
\item \textbf{Medical natural language inference tasks}, including CHIP-STS \cite{CHIP-STS}, KUAKE-QQR, KUAKE-IR, and KUAKE-QTR tasks, which asks a model to determine the semantic relations between a pair of medical queries or a query-document pair. 
\item \textbf{Symptom status understanding for medical dialogues}. Medical dialogues, like medical consultations, are centered around the patients' symptoms. However, not every symptom mentioned in the dialogues is related to patients or reflects patients' current medical status. Thus, to gain a deep understanding of the doctor-patient dialogues, a model must be able to extract the symptoms and their statuses. This cohort includes (a) IMCS-V2-SR and (b) CHIP-MDCFNPC for extracting clinical findings and their status. This cohort of tasks is based on the medical dialogue datasets IMCS-V2 \cite{Chen2022ABF} and CHIP-MDCFNPC \cite{MDCFNPC}. 
\item \textbf{Medical content generation}, which includes two generation tasks based on medical dialogues between patients and doctors: (a) IMCS-V2-MRG \cite{Chen2022ABF}, which asks LLMs to summarize the medical dialogues. (b) MedDG \cite{Liu2020MedDGAL}, which asks LLMs to act like a doctor and respond to the patient's queries in a dialogue. 
\end{itemize}

\subsection{Prompt collection} 

We employ both manual efforts and ChatGPT to collect prompt templates for PromptCBLUE. The prompt templates mainly contain task instructions describing what we want the LLMs to provide, given the text input. Firstly, each of the three annotators who are graduate students majoring in computer science and are studying LLMs will write around two seed prompt templates manually. We ask the annotators to write the prompts as diversified as possible. The prompt templates are then reviewed by a panel of two medical experts and a senior NLP researcher to ensure their validity. If a prompt is not proper (for example, not expressing the task clearly), we will ask the annotators to modify it until the domain experts accept it. Then, we will ask the ChatGPT to rephrase each of the six seed prompts templates ten times without changing the meaning or changing the placeholders in the templates. Then, the generated templates will be reviewed by the same panel of experts, and only the templates passing the reviews will be added to the template pool. After the prompt collection process, there are a total of 320 prompt templates in PromptCBLUE.\footnote{The manually written prompt templates and the augmented template sets are open-sourced at our code repository, https://github.com/michael-wzhu/PromptCBLUE. }

\subsection{Response format} 

Note that LLMs can only generate token sequences to represent their answers to the queries. We have to transform the structured outputs of the original CBLUE tasks into natural language sequences. In the Appendix, we present the target output formats for each task under PromptCBLUE. 

Recently, \cite{Wei2022ChainOT,Wang2022SelfConsistencyIC} show that LLMs' reasoning or task-solving ability can be improved if LLMs are asked to complete the task step-by-step or reason step-by-step, that is, chain-of-thought (COT) prompting. In this task, inspired by \cite{Wei2022ChainOT}, we will explore the idea of COT in the design of prompt and response formats for complex medical information extraction tasks. Take CMeIE as an example. With COT, in the prompt, we will ask the LLM to determine the existing relations in the given sentence and then extract the triples for each relation. Accordingly, in the response, first, the relations will be presented, and then the extracted triples will be organized into relation groups. Without COT, the prompt will not instruct the LLM to solve the task step-by-step, and in the response, the triples will be generated one by one. In this work, unless stated otherwise, we will use the prompt and response formats with COT for the CMeEE-V2, CMeIE, CHIP-CDEE, and IMCS-V2-NER tasks. We will use ablation studies to demonstrate that COT benefits in-context learning of LLMs like ChatGPT and fine-tuned open-sourced LLMs.

\subsection{Sample format} 

In PromptCBLUE, all the samples from different tasks are organized in the following data structure: 
\begin{verbatim}
{
  "input": str,
  "target": str,
  "answer_choices": Union[list, Nonetype],
  "sample_id": str,
  "task_type": str,
  "task_dataset": str,
}
\end{verbatim}
Here, \emph{input} is the prompt sequence, \emph{target} is the response we want the model to output, or at least the model should generate responses that contain the answers to the prompt. The other four keys are auxiliary, and the LLMs will not use them. \emph{sample\_id} is the sample index. \emph{answer\_choices} is the label option allowed by the prompt. The value for this key will be \emph{None} if the task does not have a predefined label set. \emph{task\_dataset} specifies the task name in the original CBLUE benchmark, and \emph{task\_type} is the original CBLUE task type.

\subsection{Dataset splits} 

Note that after filling in the CBLUE data samples in the prompt templates, the train/dev/test sets for some tasks will be quite large. Considering that LLMs require high computation resources to fine-tune and have large latency for token sequence generation, we limit the training samples of each task to 3000 to 5000 and the dev/test set to 400. We first fill each prompt template with the samples to construct a large test sample pool and randomly select the prompt-response pairs via uniform sampling.\footnote{Note that the training set is provided just as a reference for participants in the Tianchi competition since some of the original CBLUE tasks have large training sets.} In Table \ref{tab:stats}, we present the dataset statistics for the PromptCBLUE dataset.

\begin{table}[t]
\caption{Dataset statistics for the PromptCBLUE benchmark. }
\label{tab:stats}
\begin{center}
\begin{tabular}{c|ccc}
\hline
\textbf{Task}   &   \#Train/\#dev/\#test  &  Prompt length    &   target length      \\
\hline
\multicolumn{4}{c}{\textbf{\emph{Sub-tasks}}}  \\
\hline

CMeEE-V2  &  5000/400/400    &   107.88   &   54.03   \\
CMeIE   &  5000/400/400    &    293.72   &  135.51  \\
CHIP-CDEE   &  3000/400/400   &  142.61   &  180.93    \\
CHIP-CDN   &  5000/400/400    &   281.79   &  10.37    \\
CHIP-CTC   &   6600/704/704    &   214.61   &    3.81   \\
CHIP-STS   &   5000/400/400    &   66.26    &   2.0   \\
KUAKE-QIC   &   5500/440/440    &  81.58   & 
   4.09   \\
KUAKE-QTR   &  5000/400/400    &    96.38   & 
  7.23   \\
KUAKE-QQR   &  5000/400/400    &   89.38    &  7.61   \\

KUAKE-IR   &    5000/400/400    &     203.33   &    2.78    \\
CHIP-MDCFNPC  &   5000/400/400    &   744.99  
 &   67.67   \\
IMCS-V2-SR  &   5000/400/400    &   137.13   &    36.33   \\
IMCS-V2-NER   &   5000/400/400    &  61.66  & 
  23.65   \\
IMCS-V2-DAC    &   5000/512/512    &     371.62   &     8.56   \\
IMCS-V2-MRG   &    3000/400/400     &    821.1   &  105.08   \\
MedDG    &   5000/400/400    &   194.75   &  27.71  \\ 
Text2DT  &   1500/400/400   &   465.10   &  392.47                \\
CMedCausal  &     3000/400/400     &  423.68    &  272.25         \\
\hline
\multicolumn{4}{c}{\textbf{\emph{Total}}}  \\
\hline
PromptCBLUE   &    87100/8456/8456     &    265.22   &  71.10    \\
\hline

\end{tabular}
\end{center}
\end{table}

\textbf{Quality Checking} \quad The quality check of our data is conducted in the following three aspects:
\begin{itemize}
\item Ensuring the quality of the prompt templates with the help of the expert panel, as described above.
\item Checking the quality of the CBLUE benchmark. During the development of the PromptCBLUE benchmark, we are also helping the CBLUE benchmark to improve the annotations. For example, we have found that the original QQR task has an inappropriate label set and asked the CBLUE organizers to re-annotate this task. 
\item Random sampling. To ensure the data quality, we sampled 5\% or 200 of the samples from each task of PromptCBLUE, and each sample was examined by a group of annotators from the medical field. Finally, we identify an average of 0.9\% mislabeling rate. Based on the evaluation results in the next section, such an error rate will not significantly impact the overall evaluation accuracy.
    
\end{itemize}

\section{Participating Teams and Methods}

\subsection{Participating Teams}

This shared task is held in CHIP-2023 Conference as Shared Task 1. We held this shared task with the help the Tianchi Team, and all the logistics are handled by the Tianchi platform.\footnote{The PEFT track is at \url{https://tianchi.aliyun.com/competition/entrance/532132}, and the ICL track is held at \url{https://tianchi.aliyun.com/competition/entrance/532131}. } In the first round of the shared task, 362 teams participated in the PEFT track, and 238 teams participated in the ICL track. In the second round, 31 teams submitted predictions for the PEFT track, and 17 teams submitted predictions for the ICL track.

\subsection{Wining teams}

For both tracks, we rank the teams by the average score of all the 18 sub-tasks in the PromptCBLUE benchmark. In accordance with the regulations of the shared tasks in CHIP-2023, the top-3 teams of each track are eligible for the awards and certificates issued by the CHIP-2023 committee. And these teams are invited to submit papers to share their techniques and experiences to the community.\footnote{The third place of the PEFT track is kept empty since the corresponding team does not submit their materials for review. } 

In the table \ref{tab:all_test_results} below, we have listed the top-3 teams from each track. We can see that there is a clear gap between the two tracks with respect to the average score, showing that the open-sourced LLMs can not perform as well as fine-tuning under the ICL setting.

\begin{table}
\caption{The winning teams and their test results of the PromptCBLUE shared task. }\label{tab:all_test_results}
\centering
\begin{tabular}{p{1.5cm}p{1.5cm}p{5.0cm}p{2.0cm}}
\hline
Team name &   Rank   &  Institution   &  	Avg Score  \\
\hline
\multicolumn{4}{c}{\textbf{Winners of the PEFT track}}   \\
\hline

pt.boys   &   1    &  \makecell[l]{Huimei Healthcare Management \\Services (惠每科技) }    &   71.38   \\
练习一下   &   2    &   \makecell[l]{苏州大学 (University of Suzhou)\\~\& 北京邮电大学 (Beijing \\University of Posts and \\Telecommunications) }    &   68.09    \\ 

\hline
\multicolumn{4}{c}{\textbf{Winners of the ICL track}}   \\
\hline

紫丁香队    &   1    &  \makecell[l]{哈尔滨工业大学 (Harbin Institute \\of Technology)}   &   40.27   \\
ECNU-LLM    &   2    &   \makecell[l]{华东师范大学 (East China \\Normal University) }  &    39.24    \\
IMI1214   &   3    &   \makecell[l]{上海理工大学 (University of \\Shanghai for Science and \\Technology)~\& 南京大学 (Nanjing \\University) }    &   36.80   \\
			
\hline
\end{tabular}
\end{table}

\subsection{Methods of the PEFT track}

In this subsection, we will analyze the methods adopted by the top ranked teams in the PEFT track. 

\noindent\textbf{Pre-trained backbones} \quad In this shared task, to ensure fair comparisons and not to turn the whole task to a search of stronger LLM backbone, we only allow the following open-sourced Chinese LLM backbone models: (a) ChatGLM-6B-2\footnote{\url{https://huggingface.co/THUDM/chatglm2-6b}.}; (2) Baichuan-13B-Chat\footnote{\url{https://huggingface.co/baichuan-inc/Baichuan-13B-Chat}.}; (3) Chinese-LlaMA2-7B-chat\footnote{\url{https://huggingface.co/michaelwzhu/Chinese-LlaMA2-chat-7B-sft-v0.3}.}; (4) Chinese-LlaMA2-13B-chat\footnote{\url{https://huggingface.co/michaelwzhu/Chinese-LlaMA2-13B-chat}.}. From the results of the winning teams in the PEFT track, we can see that: (a) the 13B models perform better 7B models under PEFT tuning; (b) the Baichuan-13B-Chat model obtains the strongest performance.

\noindent\textbf{Data processing and augmentation} \quad Note that our PromptCBLUE benchmark comes with a training set, and naturally it would enhance the model performance if we conduct certain data augmentation operations on the training set. The following are the data augmentation operations investigated by the winning teams: 
\begin{itemize}
\item Note that our PromptCBLUE benchmark comes with a training set containing at most 5k to 6.6k samples for each sub-task. However, some tasks from the original CBLUE benchmark have a large training set. Thus, we can augment the training set with the original CBLUE ones. For example, Team pt.boys augments the training sets of the CHIP-CDEE, CMeEE-V2, CHIP-CDN, CMeIE-V2, IMCS-V2-NER, CHIP-MDCFNPC, CHIP-CTC, CHIP-STS and CMedCausal sub-tasks with the CBLUE data by employing the prompt templates open-sourced by us\footnote{The prompt templates of the PromptCBLUE benchmark are open-sourced at \url{https://github.com/michael-wzhu/PromptCBLUE/blob/main/src/data/templates\_augment.json}. }. They augment each of the above mentioned sub-task's training sample size to 16000. To ensure task samples' balance, they also up-sample the other sub-tasks' sample sizes to 8000.
\item The previous literature \cite{zhu-etal-2021-discovering,Feng2021ASO,Zhang2023LECOIE,Zhang2023NAGNERAU} shows that augmenting training samples by randomly masking the contents of the input sentence can improve the robustness of the trained models. Team pt.boys randomly replace the token in the sentence with the <unk> token, for the sentence classification tasks in PromptCBLUE (IMCS-V2-DAC, CHIP-CTC, CHIP-STS, KUAKE-IR, KUAKE-QIC, KUAKE-QQR, and KUAKE-QTR) 
\end{itemize}

\noindent\textbf{The parameter-efficient fine-tuning methods} \quad The PEFT methods rises ffollowing the rise of pre-trained language models, especially when the research field made efforts to models with larger scales. There are a variety of PEFT methods in the pre-LLM era. However, since the rise of LLMs, especially the open-sourced LLMs, LoRA becomes the most popular PEFT method. We believe that the popularity of LoRA comes from the following three reasons: (a) LoRA is a reparameterization of the original Transformer weights, thus it can be merged to the model backbone and introduce no additional latency; (b) LoRA has a good theoretical basis: it comes from the idea of intrinsic space, and the fact that fine-tuning a well pretrained model is in essence low-rank. (c) open-source implementations. LoRA is implemented by many open-sourced code repositories like the Huggingface PEFT package\footnote{\url{https://github.com/huggingface/peft}.}, thus it is very convenient to use. In the PEFT track, all the winning teams use the LoRA method to fine-tune the LLM backbones. 

\noindent\textbf{Training techniques} \quad In order to fine-tune the model properly, a few issues should be addressed. First, despite the fact that most the LLM backbone is freezed and we only update the LoRA parameters, the LLMs still requires a large GPU memory, which may be difficult to obtain for academia. Thus, Team "联系一下" employ the QLoRA method \cite{qlora} to reduce the memory requirements during fine-tuning. QLoRA proposes a novel 4-bit quantization method that proves to have minimum effects for the model fine-tuning. In addition, Flash Attention is also applied to reduce the memory comsumption when dealing with long sequences. Second, since we are dealing with a multi-task prompt tuning task, many teams pay attention to balance the sample sizes of the subtasks, so that the fine-tuned LLM will perform propoerly on each subtask.

\subsection{Methods of the ICL track}

In this subsection, we will analyze the winning teams' techniques in the ICL track.  

\noindent\textbf{Pre-trained backbones} \quad From the results of the winning teams in the PEFT track, we can see that: (a) similar to the PEFT track, the 13B models perform better 7B models in ICL; (b) Chinese-LlaMA2-13B-chat helps to achieves the best ICL results, showing that Baichuan-13B-Chat is not dominating in every aspects.

\noindent\textbf{Demonstration selection} \quad The demonstration selection method is the core of the ICL track. When the LLM is a blackbox and can not be fine-tuned, such as model APIs, ICL capabilities play an important role for unseen or emergent tasks. All three winning teams have used the similarity based demonstration selection. That is, for a test prompt, a sentence embedding model is employed to retrieve the most similar prompts in the training set. When computing a similarity score between the test prompt and training sample, different approaches can be employed: (a) traditional methods like BM25 \cite{Robertson2009ThePR}. (b) semantic representation, which relies on a model to transformer the sequences to hidden vectors. Team "紫丁香队" used both methods, and the other two teams mainly rely on the semantic models. Three teams apply different sentence embdding models. For example, Team ECNU-LLM applied the BGE base model \cite{bge_embedding} for semantic representations and semantic retrieval. When the similar samples are retrieved, there are many approaches to determine the final demonstration combinations:  
\begin{itemize}
\item The greedy approach. The retrieved top 3-10 training samples are used as demonstrations. 
\item Knapsack based demonstration selection. Since we are using the LLMs for inference, we have to consider the maximum length it can handle given our GPU environment. Thus, when we have a pre-defined maximum length, choosing the combination of demonstrations becomes a classic knapsack problem if we consider the similarity score of a demonstration as the value, and the sample's length as the item weight. Team "紫丁香队" has investigated this approach and find that this strategy is better than the greedy method on four of the subtasks.
\end{itemize}

\section{Conclusion}

In this article, we review the PromptCBLUE benchmark, the first large-scale Chinese medical prompt tuning benchmark. Then, we given an overview of the PromptCBLUE shared task in the CHIP-2023 conference. The shared task is a huge success, attracting participants from both the industry and academia. Then, we analyze the winning methods of the two tracks. Different techniques are investigated in the shared task to fully uncover the limit of the Chinese LLMs. 

\section{Acknowledgements}

This work was supported by NSFC grants (No. 61972155 and 62136002) and National Key R\&D Program of China (No. 2021YFC3340700), and Shanghai Trusted Industry Internet Software Collaborative Innovation Center.

\bibliographystyle{splncs04}
\bibliography{mybib}

\end{CJK*}

\end{document}